\documentclass[twoside]{article}

\usepackage[accepted]{aistats2026}

\usepackage[round]{natbib}
\renewcommand{\bibname}{References}
\renewcommand{\bibsection}{\subsubsection*{\bibname}}

\usepackage[utf8]{inputenc} 
\usepackage[T1]{fontenc}
\usepackage{hyperref}  
\usepackage{url}  
\usepackage{booktabs} 
\usepackage{multirow}
\renewcommand{\arraystretch}{0.95} 
\newcommand{\std}[1]{{\scriptsize$\pm$#1}} 
\usepackage{amsfonts}  
\usepackage{nicefrac} 
\usepackage{microtype} 
\usepackage[table]{xcolor}
\usepackage{colortbl}
\usepackage{enumitem}
\usepackage{graphicx}
\usepackage[noabbrev]{cleveref}
\usepackage{makecell} 

\usepackage{amsmath,amsfonts,amsthm,bm}
\usepackage{yhmath}
\usepackage{derivative}
\usepackage{algorithm}
\usepackage{algorithmic}
\usepackage{graphicx}

\def\eqref#1{equation~\ref{#1}}

\def\1{\bm{1}}

\theoremstyle{plain}
\newtheorem{proposition}{Proposition}

\theoremstyle{definition}
\newtheorem{definition}{Definition}

\theoremstyle{remark}

\newtheoremstyle{assump}
  {3pt}{3pt} 
  {\itshape}{}  
  {\bfseries}{} 
  {0.5em}{}
\theoremstyle{assump}

\newcommand{\ptarget}{p_{\rm{target}}}
\newcommand{\pprior}{p_{\rm{prior}}}

\newcommand{\deriv}{{\mathrm{d}}}

\newcommand{\backvec}[1]{\reflectbox{$\vec{\reflectbox{$#1$}}$}}

\def\rvw{{\mathbf{w}}}
\def\rvx{{\mathbf{x}}}

\newcommand{\veps}{\boldsymbol{\varepsilon}}

\DeclareMathAlphabet{\mathsfit}{\encodingdefault}{\sfdefault}{m}{sl}
\SetMathAlphabet{\mathsfit}{bold}{\encodingdefault}{\sfdefault}{bx}{n}

\def\sP{{\mathbb{P}}}

\def\sR{{\mathbb{R}}}

\newcommand{\E}{\mathbb{E}}

\newcommand{\R}{\mathbb{R}}

\newcommand{\KL}{D_{\mathrm{KL}}}

\begin{document}

\twocolumn[
\runningtitle{One-Step Diffusion Samplers (OSDS)}
\aistatstitle{One-Step Diffusion Samplers \\ via Self-Distillation and Deterministic Flow}

\aistatsauthor{%
  Pascal Jutras-Dubé \And 
  Jiaru Zhang \And
  Ziran Wang \And
  Ruqi Zhang 
}

\aistatsaddress{%
  Purdue University 
  \And   Purdue University
  \And   Purdue University 
  \And   Purdue University 
}
]

\begin{abstract}
Sampling from unnormalized target distributions is a fundamental yet challenging task in machine learning and statistics. 
Existing sampling algorithms typically require many iterative steps to produce high-quality samples, leading to high computational costs. 
We introduce \emph{one-step diffusion samplers}, which learn a step-conditioned ODE such that a single large step reproduces the trajectory of many small ones via a state-space consistency loss.
We further show that standard ELBO estimates in diffusion samplers degrade in the few-step regime because common discrete integrators yield mismatched forward/backward transition kernels.
Motivated by this analysis, we derive a deterministic-flow (DF) importance weight for ELBO estimation without a backward kernel. 
To calibrate DF, we introduce a volume-consistency regularization that aligns the accumulated volume change along the flow across step resolutions.
Our proposed sampler therefore achieves both fast sampling and stable evidence estimate in only one or a few steps.
Across challenging synthetic and Bayesian benchmarks, it achieves competitive sample quality with orders‑of‑magnitude fewer network evaluations while maintaining robust ELBO estimates.
\end{abstract}

\makeatletter
\begingroup
    \footnotesize{
    \rule{0.8in}{0.4pt}\\[0.25ex]
    Correspondence to: Pascal Jutras-Dub\'e (\nolinkurl{pjutrasd@purdue.edu}),
    Jiaru Zhang (\nolinkurl{jiaru@purdue.edu}), and
    Ruqi Zhang (\nolinkurl{ruqiz@purdue.edu}).
    Code is available at: \url{https://github.com/PascalJD/one-step-diffusion-samplers}
}
\endgroup
\makeatother

\section{INTRODUCTION}

Sampling from densities of the form 
\begin{equation}
    \ptarget = \frac \rho Z, \quad \text{with} \quad Z = \int_{\mathbb R^D} \rho(x)\deriv x 
\end{equation}
with $\rho$ evaluable pointwise but $Z$ intractable, is a central problem in machine learning \citep{hernandez2015probabilistic,neal1995bayesian} and statistics~\citep{andrieu2003mcmc,neal2001annealed}, and has applications in scientific fields like physics \citep{albergo2019flow, No2019BoltzmannGS,wu2019solving}, chemistry \citep{frenkel2002molecularsimulation, holdijk2024ocmolecule,hollingsworth2018moleculardynamics}, and many other fields involving probabilistic models. 

Many established sampling algorithms are inherently \emph{iterative}, with the accuracy of the final samples depending heavily on the number of steps. 
Markov chain Monte Carlo (MCMC) methods converge asymptotically but require long chains in practice \citep{mackay2003mcmcbook, robert1995convergencemcmc}, while recent diffusion-based samplers \citep{berner2024dis,vargas2023dds,zhang2022pis} guarantee finite-time convergence but still rely on hundreds of discretization steps, limiting their use in large models and resource-limited scenarios.

In this paper, we develop Self-Distilled One-Step Diffusion Samplers (OSDS) that preserve sample quality and deliver reliable evidence estimates. 
Our starting point is self‑distillation by state consistency: a step‑conditioned control is trained so that a single large probability‑flow (PF) ODE step reproduces the composition of many small steps, enabling one/few‑step transport in state space.

At the same time, we uncover a critical limitation: state consistency alone does not yield a principled ELBO. 
In diffusion samplers, ELBOs are typically computed from path‑space likelihood ratio \citep{richter2024improved,vargas2024cmcd}.
In the few‑step regime, common discretizations are time‑asymmetric, so the likelihood ratio becomes unstable and ELBO estimates collapse, even when samples look accurate.

To address this, we introduce a deterministic‑flow (DF) importance weight. 
Prior samples are pushed forward by the learned PF ODE, and a log‑Jacobian is accumulated along the flow; the resulting DF weight yields stable, accurate ELBOs in the one/few‑step regime.

OSDS is trained by jointly optimizing: (i) state consistency to match a large step with the composition of small PF steps, and (ii) volume consistency to match the accumulated log‑Jacobian across resolutions, ensuring geometric fidelity of the flow.
The resulting deterministic map supports fast sampling and evidence estimation with the same small step budget.
Our contributions can be summarized as follows:
\begin{itemize}[leftmargin=*,itemsep=2pt, topsep=0pt]
    \item We show that RND path‑space weights break in the few‑step regime because common discrete integrators are time‑asymmetric, inducing a forward-backward kernel mismatch and collapsing ELBO.
    \item We propose Self-Distilled One-Step Diffusion Samplers (OSDS), which learn a step-conditioned shortcut PF ODE via state and volume consistency, and derive a deterministic‑flow importance weight from PF ODE change‑of‑variables that remains stable at one/few steps. To the best of our knowledge, this is the \emph{first} sampler that achieves both high-quality sample generation and accurate statistical estimation in one/few steps.
    \item Across a wide range of challenging synthetic and Bayesian benchmarks, OSDS delivers competitive sample quality with orders‑of‑magnitude fewer network evaluations, while maintaining robust evidence estimation in the few‑step regime.
\end{itemize}

\section{RELATED WORK}\label{sec:related-work}
\paragraph{Markov chain Monte Carlo samplers.}
Classical Monte Carlo samplers construct a Markov chain whose stationary distribution matches the target \citep{brooks2012handbookmcmc}.
Prominent examples include Metropolis–Hastings \citep{hastings1970monte,metropolis1953equation}, Gibbs sampling \citep{geman1984gibbs}, and overdamped Langevin methods \citep{parisi1981langevin,rossky1978langevin}.
By exploiting geometry, Hamiltonian Monte Carlo improves exploration \citep{brooks2012handbookmcmc,duance1987hmc,mackay2003mcmcbook}, and scalability is addressed by stochastic‐gradient variants \citep{welling2011langevin,chen2014stochastic,zhang2019cyclical,zhang2020amagold}.
Despite these advances, MCMC requires many transitions to produce high‐quality samples, and convergence is only asymptotically guaranteed.

\paragraph{Normalizing flows.}
Normalizing flows (NFs) learn invertible maps that transform a simple base distribution into the target.
Continuous‐time variants (CNFs) evolve samples by integrating an ODE with an iterative solver \citep{chen2018neuralode,grathwohl2019ffjord}.
In the unnormalized‐density setting considered here, flows are typically trained through variational objectives or combined with auxiliary procedures (e.g., annealing or SMC) \citep{wu2020snf, Arbel2021aft, matthews2022craft}.

\paragraph{Generative flow networks.}
GFlowNets (GFNs) amortize sampling from intractable distributions by learning policies that construct samples via multi‐step trajectories \citep{bengio2021gflownets,bengio2023foundations}.
Recent theory extends GFNs to continuous or hybrid spaces, connecting them to diffusion and flow matching \citep{lahlou2023continuousgflownets, zhang2024dgfs, berner2025gflow}.
While GFNs amortize training‐time exploration, their inference remains iterative because a trajectory must be generated step by~step.

\paragraph{Bridge samplers.}
Schr\"odinger bridge (SB) methods cast sampling as an entropic optimal transport problem, learning a stochastic process that bridges a prior to the target; see, e.g., SB samplers \citep{bernton2019sbs} and diffusion Schr\"odinger bridges (DSB) \citep{debortoli2021dsb}.
Underdamped diffusion bridges (UDB) broaden this view to degenerate noise and exploit favorable convergence/numerical properties of underdamped dynamics \citep{blessing2025udb}.
Bridge‐based samplers learn or estimate dynamics and thus require discrete-stepping SDE integration.

\paragraph{Diffusion‐based samplers.}
A complementary control-theoretic perspective learns drifts that transport an easy prior to the target by solving a (stochastic) optimal control problem \citep{berner2024dis,richter2021phd,richter2024improved,zhang2022pis}.
This viewpoint underlies many recent diffusion samplers \citep{doucet2022dais,geffner2023langevin,vargas2023dds} and refinements (e.g., resampling \citep{chen2025SCLD}, Hamiltonian couplings \citep{blessing2025udb}, particle methods \citep{phillips2024particle}, PDE-based evolution \citep{sun2024pinn}, simulation-free training \citep{akhound2024idem}).
While these methods amortize some cost in training, inference still integrates dynamics over many steps.

\paragraph{Acceleration techniques for diffusion models.}
A parallel line of work reduces the number of reverse-time steps in data-driven diffusion models. 
Consistency models enable one-step generation by distilling a pretrained diffusion model \citep{song2023consistency} or can be trained from scratch \citep{song2023improved, lu2025simplifying}. 
Progressive distillation repeatedly halves the number of solver steps while preserving sample quality \citep{salimans2022progdist}. 
More recently, shortcut models train a single network to “skip” multiple diffusion steps in one shot \citep{frans2025shortcut}. 
Orthogonally, solver-level accelerations such as high-order solvers \citep{jolicoeurmartineau2021gotta, lu2022dpm, karras2022edm} and straightened transport paths \citep{liu2023flowstraight} permit large step sizes at inference. 
These accelerators assume access to data from the target distribution, are not designed to incorporate only pointwise evaluations of an unnormalized density, and typically do not provide estimates of the log-partition function.

\paragraph{Summary.}
Prior samplers rely on multi-step trajectories or time integration at inference. 
Recent accelerators for diffusion models reduce that step count, but they are data-driven and do not natively handle unnormalized targets or yield $\log Z$ estimates. 
By contrast, our method fully amortizes exploration during training and produces single-step samples at test time from unnormalized densities given only pointwise access to~$\rho$, while supporting normalization estimates.

\section{PRELIMINARIES}\label{sec:preliminaries}
\subsection{Diffusion Samplers and SDEs}
Diffusion samplers aim to draw samples from a complex target density \(\ptarget = \rho / Z\) by transporting them from a simpler prior density \(\pprior\). 
This is modeled as a generative Stochastic Differential Equation (SDE) in \(\sR^D\) forward in time \(t\in[0, T]\):
\begin{equation}\label{eq:generative-sde}
    \deriv \rvx_t = \mu(\rvx_t, t)\deriv t + \sigma(t) \deriv \rvw_t, \quad \rvx_0 \sim \pprior, 
\end{equation}
where \(\mu(\rvx_t, t) \in \mathcal{U} \subset C(\sR^D \times [0,T], \sR^D)\) is a learnable drift, \(\sigma(t) \in C([0,T], \sR)\) is a predefined diffusion coefficient, and \({\deriv}\rvw_t\) denotes the standard Wiener increments.
The core objective is to learn the drift $\mu$ such that this generative process becomes the time-reversal of a fixed, linear noising SDE:
\begin{equation}\label{eq:noising-sde}
    \deriv \rvx_t = \backvec f(t) \rvx_t\deriv t + \backvec\sigma(t)\deriv  \rvw_t, \quad \rvx_0 \sim \ptarget,
\end{equation}
where \(f \in  C([0,T], \sR)\) and \(\backvec f(t) = f(T - t)\) are predefined. 
It is known that this time-reversal is achieved if the generative drift satisfies 
\begin{align}
\mu = \sigma \sigma^\top \nabla_\rvx \log p_t(\rvx_t) - f, \label{eq:optimal_mu} 
\end{align}
where \(p_t(\rvx_t)\) is the density of \(\rvx_t\) \citep{Anderson1982ReversetimeDE,follmer1986time, nelson1967dynamical}.

In practice, the time-marginals \(p_t\) are intractable, which means the ideal drift in \cref{eq:optimal_mu} cannot be computed directly. 
Instead, the intractable scaled score term \(\sigma  \nabla_\rvx \log p_t\) is parameterized by a neural network expressed as a control function $u_\theta(x, t)$. 
It is trained by minimizing a divergence $D(\sP|\backvec{\sP})$ between the path measures of the generative process ($\sP$) and the noising process ($\backvec{\sP}$). 
This requires computing the likelihood ratio between the two continuous-time path measures, which is given by the following proposition, with the proof deferred to Appendix~\ref{sec:rnd}.
See also \cite[Proposition 2.2]{vargas2024cmcd}.
\begin{proposition}[Forward-backward Radon-Nikodym derivative (RND)]\label{prop:rnd}
Let $\sP$ be the path measure of the forward SDE in \cref{eq:generative-sde} with control $u$, and let $\backvec{\sP}$ be the path measure of the reverse-time SDE in \cref{eq:noising-sde} with terminal density $\ptarget$.

Then for any path $\rvx$ we have
\begin{align}\label{eq:rnd}
\begin{aligned}
        \log\frac{\deriv {\sP} }{\deriv \backvec{\sP}}(\rvx) &= \frac{1}{2}\int_0^T\left(\|\sigma f\|^2 - \|\sigma f + u_\theta\|^2\right)(\rvx_t, t)\deriv t \\ &+ \int_0^T\left(\frac{u_\theta}{\sigma}\right)(\rvx_t, t)\cdot\deriv \rvx_t + \log\frac{\pprior(\rvx_0)}{\ptarget(\rvx_T)}.
        \end{aligned}
\end{align}
\end{proposition}

Once trained, the optimized control \(u_\theta\) allows generation of samples from \(\ptarget\) through forward simulations of \cref{eq:generative-sde}. 
In practice, this continuous-time process must be discretized into finite steps \(0 = t_1 < t_2 < \dots < t_N = T\), introducing a trade-off between computational cost and accuracy.
\subsection{The FB-RND Framework for Evidence Estimation}
Another key task in diffusion samplers beyond sample generation is estimating the model evidence, \(\log Z\), which is crucial for model comparison and selection. Diffusion samplers are able to provide this estimate. 
The theoretical foundation for this capability is the discrete Forward-Backward Radon-Nikodym Derivative (FB-RND) framework, which provides a valid Evidence Lower Bound (ELBO) on \(\log Z\). This is formally stated in the following proposition:
\begin{proposition}[Finite-step forward-backward bound]\label{prop:disc-fb}
For every discrete trajectory $\rvx_{0:N}$ simulated under the forward discretization of \cref{eq:generative-sde},
\begin{align}\label{eq:disc-fb-rnd-main}
\begin{aligned}
& \log\frac{\deriv \sP}{\deriv \backvec{\sP}}(\rvx_{0:N}) \\
=& \log\frac{\pprior(\rvx_0)}{\ptarget(\rvx_N)} + \sum_{n=0}^{N-1}\log\frac{p_{n+1|n}(\rvx_{n+1} | \rvx_n)}{p_{n|n+1}(\rvx_n | \rvx_{n+1})}.
\end{aligned}
\end{align}
Taking expectation with respect to the forward discrete path measure yields the lower bound
\begin{align}\label{eq:disc-fb-bound-main}
\begin{aligned}
\E_{\sP}\Big[\log\rho(\rvx_N) - \log\pprior(\rvx_0) + \\ \sum_{n=0}^{N-1}\log\frac{p_{n+1|n}(\rvx_{n+1} | \rvx_n)}{p_{n|n+1}(\rvx_n | \rvx_{n+1})}\Big] \le \log Z.
\end{aligned}
\end{align}
See Appendix \ref{sec:disc-fb} for a proof and a Monte Carlo estimator of \(\log Z\), and \Cref{alg:fb-rnd-bound} for pseudocode.
\end{proposition}
The bound shown by \cref{prop:disc-fb} provides an effective evaluation approach of \(\log Z\) in diffusion samplers.

\section{ONE-STEP TRANSPORT AND EVIDENCE ESTIMATION}
This section introduces Self-Distilled One-Step Diffusion Samplers (OSDS), a framework designed to achieve two complementary goals: (i) generating high-fidelity samples in a single or very few steps, and (ii) reliably estimating the partition function \(\log Z\) with the same few-step budget.
The core idea is to self-distill a diffusion sampler into a deterministic flow map induced by its PF ODE.    
OSDS is trained using two complementary consistency terms: a state-space self-distillation loss, which enforces that the transported states match across resolutions, and a volume consistency loss, which enforces that the log-volume change (log-determinant of the Jacobian) accumulated by a large step matches the sum of the composition of small steps.  
At test time, the resulting flow map is used for both single/few-step sampling and estimation of \(\log Z\) via change of variables.

\subsection{Learning One‑Step Transport via State Consistency}\label{sec:baeline}
\begin{figure*}[bt!]
\centering
\includegraphics[width=0.95\linewidth]{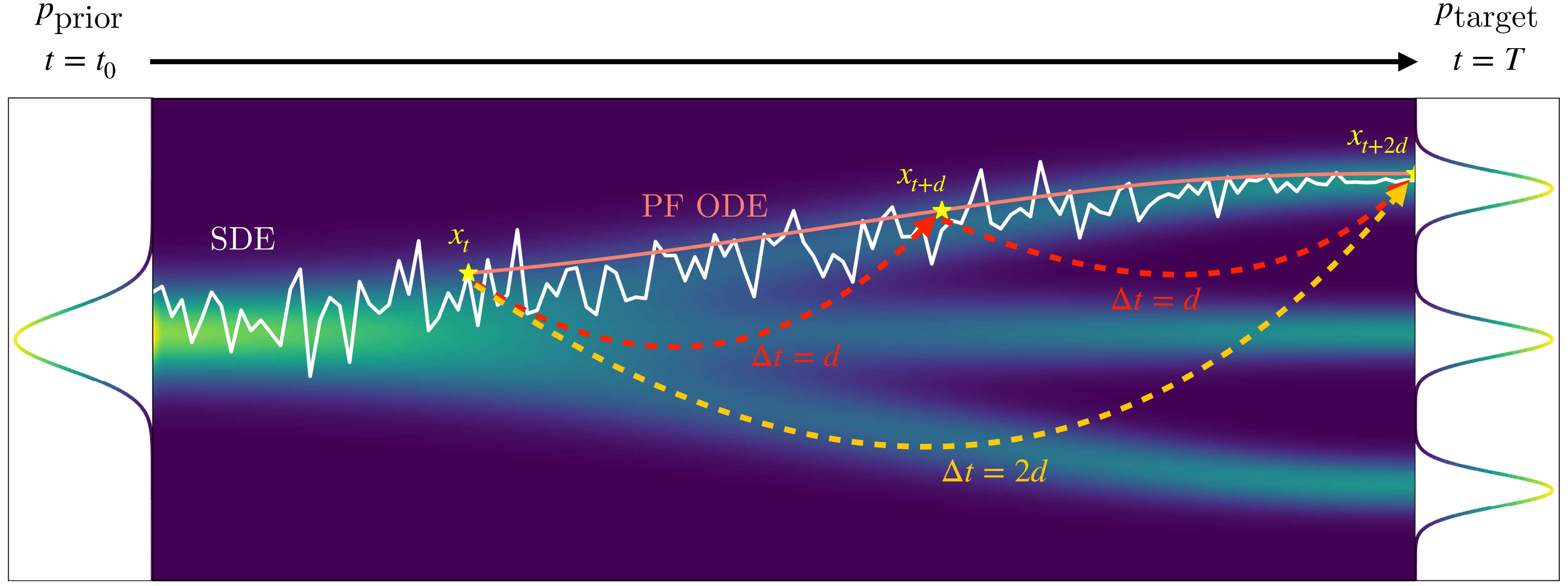}
\caption{Graphical illustration of state-space self-distillation.}
\label{fig:state-distillation}
\end{figure*}
Our self-distillation framework relies on a recursive consistency principle: a large step should behave like the composition of smaller steps. 
To anchor this recursion, we need a reliable small-step sampler.
In our setting there is no dataset to reveal the modes of \(\ptarget\); the sampler must discover them.  
A diffusion sampler does exactly this: its Brownian motion visits high-density regions, leading to coverage that a purely deterministic flow (or a likelihood-trained normalizing flow without data) would struggle to obtain.

\paragraph{RND base loss at small steps.}
At a base resolution \(d_0 = T/N_0\), we train \(u_\theta(x,t,d_0)\) as a diffusion sampler using the KL objective
\begin{equation}\label{eq:rnd-loss}
    \mathcal L_{\text{RND}} = \E\Big[\log \frac{\deriv \sP}{\deriv \backvec{\sP}}\Big].
\end{equation}
where the likelihood ratio is given by the RND \cref{prop:rnd}.
This ensures the model learns a valid diffusion sampler at fine resolution.

\paragraph{Self-distillation to larger steps.}
Let \(\Psi_\theta:\R^D\times[0,T]\times(0,T]\to\R^D\) denote a generic numerical solver that advances the state of the PF ODE of the generative SDE \cref{eq:generative-sde}:
\begin{equation}\label{eq:pf-ode}
    \deriv \rvx_t = \big(\tfrac{1}{2}\sigma u_\theta - f\big)(\rvx_t, t, d)\deriv t, \quad \rvx_0 \sim \pprior, 
\end{equation}
from time \(t\) to \(t{+}d\): \(\widehat x_{t+d} =\ \Psi_\theta(x_t;t,d)\).

We form a teacher–student pair: the teacher is the composition of two half-steps under frozen parameters \(\theta'=\operatorname{stopgrad}(\theta)\),
\begin{equation*}
    x_{t+\frac d2}=\Psi_{\theta'}(x_t;t,\tfrac d2),\qquad
    x_{t+d}=\Psi_{\theta'}(x_{t+\frac d2};t+\tfrac d2,\tfrac d2),
\end{equation*}
and the student is the single large step \(\widehat x_{t+d}=\Psi_\theta(x_t;t,d)\).
The self-distillation loss enforces
\begin{equation}\label{eq:state-consistency-loss}
    \mathcal L_{\text{state}} = \E\big[\|\widehat x_{t+d}-x_{t+d}\|^{2}\big]
\end{equation}
with $(t,d)$ sampled along powers of two from the training trajectories.

By optimizing $\mathcal L_{\text{RND}}$ and $\mathcal L_{\text{state}}$ jointly, the model learns to explore through stochastic diffusion steps and to compress this exploration into accurate large-step deterministic updates. 
\Cref{fig:state-distillation} illustrates the self-distillation procedure along the PF ODE.

\subsection{Why Path‑Space ELBO Degrades at Few Steps}
The validity of the bound proposed in \cref{prop:disc-fb} hinges on the use of a true time-reversed kernel, known as the \textit{time-adjoint} kernel \(p^{\star}_{n|n+1}\). 
Let \(\widetilde p_{n|n+1}(\cdot \mid \rvx_{n+1})\) be any surrogate backward kernel with the same support as \(p^{\star}_{n|n+1}\) (e.g., obtained by applying Euler–Maruyama to the reverse-time SDE). 
Replacing the time-adjoint kernel by \(\widetilde p_{n|n+1}\) introduces an entropy gap in the per-step forward–backward term:
\begin{align}\label{eq:entropy-gap-main}
\begin{aligned}
        &\E\big[\log p_{n+1|n}(\rvx_{n+1}| \rvx_n)-\log \widetilde p_{n|n+1}(\rvx_n| \rvx_{n+1})\big] \\
        =& -\E_{\rvx_{n+1}}\Big[\mathrm{KL}\big(p^{\star}_{n|n+1}(\cdot|\rvx_{n+1})\|\widetilde p_{n|n+1}(\cdot|\rvx_{n+1})\big)\Big]\le 0.
\end{aligned}
\end{align}
While this gap is negligible for small step sizes, it becomes dominant in the few-step regime.
For example, at \(N=1\) this single KL term can be very large because the forward EM mean and variance scale with \(\beta(t_1)\) at the right endpoint and with the control \(u\), whereas a surrogate backward that ignores these choices concentrates elsewhere. 
It catastrophically inflates the variance of the importance weights and collapses the ELBO estimates.

To further illustrate the severity of this failure, we provide a numerical example and more details in \cref{sec:em-example}. 
For a simple 1D VP SDE with a single discretization step, we explicitly calculate and compare the true backward kernel with the surrogate ``EM backward'' kernel. The results show that their variances differ by nearly 50-fold. 
This discrepancy translates to a per-step cross-entropy gap of approximately -40 nats, providing clear evidence for the collapse of the ELBO estimator.

Therefore, to achieve both fast sampling and reliable model evaluation, an alternative route to \(\log Z\) that does not rely on a fragile backward Markov kernel is required.

\subsection{Deterministic‑Flow Importance Weights}
To keep the estimator well-conditioned at one or a few steps, we complement the state-consistency with a deterministic importance sampler built from the PF ODE of the learned dynamics.
The key idea is to push forward prior samples through the single (or few) large PF step and evaluate a change-of-variables weight.

Fix a step size \(d>0\) (e.g., \(d=T\) for a full-horizon transport). 
For $t\in[0,d]$, consider the PF ODE
\begin{equation}
\label{eq:pf-ode-estimator}
\frac{\deriv \rvx_t}{\deriv t} = b_\theta(\rvx_t,t), 
\ b_\theta(x,t):=\big(\tfrac{1}{2}\sigma u_\theta - f\big)(x,t,d), 
\end{equation}
and let \(\phi_t:\R^D\to\R^D\) denote its flow map, i.e., \(\phi_0(x)=x\) and \(\frac{\deriv}{\deriv t}\phi_t(x)=b_\theta(\phi_t(x),t)\). 
The single-step deterministic transport is
\[
T(x):=\phi_d(x).
\]
We assume $b$ is sufficiently regular so that \(t\mapsto \phi_t\) is a diffeomorphic flow on \([0,d]\), see \cite[Section 6]{chen2018ctnf}. 
We will use \(x_0\sim\pprior\) and set \(y=T(x_0)=\phi_d(x_0)\).

\begin{proposition}[Deterministic-flow IS weight, \cref{sec:det-logz}]
\label{prop:det-flow-weight}
Let \(q:=T_{\#}\pprior\) be the push-forward of the prior by the flow map \(T=\phi_d\). Then for any $x_0\in\R^D$,
\begin{equation}
\label{eq:det-flow-weight-main}
w(x_0) = \frac{\rho \big(T(x_0)\big)}{q\big(T(x_0)\big)} = \rho \big(\phi_d(x_0)\big)\frac{\big|\det\nabla T(x_0)\big|}{\pprior(x_0)}.
\end{equation}
Moreover, $Z = \E_{x_0\sim\pprior}\big[w(x_0)\big]$ so
\begin{equation}
\label{eq:Z-IS-main}
\widehat Z = \frac1M\sum_{i=1}^M w\big(x_0^{(i)}\big)\;\;\text{is an unbiased estimator of }Z.
\end{equation}
\end{proposition}
For numerical stability we evaluate
\begin{equation}
\label{eq:logw-basic}
\log w(x_0)
=\log\rho(y)+\log\big|\det\nabla T(x_0)\big|-\log\pprior(x_0).
\end{equation}

\paragraph{Computing $\log|\det\nabla T|$ via the PF ODE.}
Directly forming \(\nabla T\) is expensive in high dimensions. 
For ODE flows one can compute the log-Jacobian determinant through the instantaneous change of variables identity \citep{chen2018ctnf}:
\begin{equation}
\label{eq:logdet-ode}
\frac{\deriv}{\deriv t}\log\big|\det\nabla \phi_t(x_0)\big| = \nabla \cdot b_\theta\big(\rvx_t,t\big)\ \ \text{along }\rvx_t=\phi_t(x_0).
\end{equation}
Integrating \cref{eq:logdet-ode} from \(0\) to \(d\) gives
\begin{equation}
\label{eq:logdet-integral}
\log\big|\det\nabla T(x_0)\big| = \int_{0}^{d} \nabla \cdot b_\theta\big(\rvx_t,t\big) \deriv t,\ \ \rvx_t=\phi_t(x_0).
\end{equation}
Hence one can evolve, in tandem with \(\rvx_t\), a scalar volume accumulator \(\ell_t\) solving the ODE
\begin{equation}
\label{eq:ell-ode}
\dot{\ell}_t = \nabla\cdot b_\theta(\rvx_t,t),\ \ \ell_0=0,\ \ \Rightarrow\ \ \log\big|\det\nabla T(x_0)\big|=\ell_d.
\end{equation}

\paragraph{Practical computation.}
Combining \cref{eq:logw-basic} and \cref{eq:ell-ode}, the single-sample log-weight is
\begin{equation}
\label{eq:logw-final}
\log w(x_0) = \log\rho \big(\rvx_d\big) + \ell_d - \log\pprior(x_0),
\end{equation}
with $\dot{\rvx}_t=b_\theta(\rvx_t,t),\ \rvx_0=x_0$ and $\dot{\ell}_t=\nabla\cdot b_\theta(\rvx_t,t),\ \ell_0=0$.
In high dimensions the divergence \(\nabla\cdot b\) can be obtained without forming Jacobians explicitly using standard stochastic trace estimators (e.g., Hutchinson, see \cref{sec:hutchison} and \citep{grathwohl2019ffjord}). 
The ODE pair \((\rvx_t,\ell_t)\) can then be integrated with any accurate solver; the resulting \(w(x_0)\) is plugged into \cref{eq:Z-IS-main}.

Unlike the discrete forward/backward likelihood ratio \cref{eq:disc-fb-rnd-main}, the deterministic weight does not require a reverse one‑step kernel; it only needs the forward PF‑ODE flow’s Jacobian determinant, for which continuous‑time change‑of‑variables is exact. 

\paragraph{Volume consistency for accurate log‑Jacobians.}
The accuracy of this deterministic estimator hinges on the reliability of the computed log-Jacobian. 
The RND base loss $\mathcal{L}_{RND}$ and the self-distillation loss  \(\mathcal{L}_\text{state}\) do not calibrate how densities transform, with $\mathcal{L}_{\text{state}}$ being a simple MSE on positions and insensitive to the geometric properties of the map.
Therefore, two maps can land at the same state yet induce different local volume changes. To ensure the learned flow is geometrically consistent across different step sizes, we introduce an additional Volume Consistency Loss \(\mathcal{L}_\text{vol}\).

Based on the principle that log-volumes are additive under composition, we extend the teacher-student framework to the log-Jacobian determinants. The teacher's accumulated log-volume is the sum of two half-step volumes:
\begin{equation}
    \text{vol}_\text{teacher} = \text{vol}\big(\Psi_{\theta'}(x_t,t,\tfrac d2)\big)+ \text{vol}\big(\Psi_{\theta'}(x_{t+\frac d2},t+\tfrac d2,\tfrac d2)\big),
\end{equation}
where \(\text{vol}(\cdot)\) denotes the log-Jacobian of a map, computed via \cref{eq:logdet-integral}. The student's log-volume, \(\text{vol}_\text{student} = \text{vol}(\Psi_\theta(x_t;t,d))\), is then trained to match the teacher's via a mean-squared error objective:
\begin{equation}\label{eq:vol-loss}
\mathcal L_{\text{vol}} =\E\Big[ \big(\text{vol}_{\text{student}}-\text{vol}_{\text{teacher}}\big)^2\Big].
\end{equation}
This loss acts as a regularizer, penalizing mismatches in the accumulated log-Jacobian. 
This discourages exploding or vanishing local volumes, improves the numerical stability of the PF ODE integration, and yields better-conditioned deterministic weights for the estimator. 

\subsection{The Final OSDS Framework}\label{sec:OSDS-framework}
By jointly optimizing all three objectives in our self-distillation framework, we present our final Self-Distilled One-Step Diffusion Sampler (OSDS) framework. It is trained by jointly minimizing a single, composite objective:
\begin{equation}\label{eq:final-loss}
    \mathcal{L}_\text{OSDS} = \mathcal{L}_\text{RND} + \mathcal{L}_\text{state} + \lambda_\text{vol}\mathcal{L}_\text{vol},
\end{equation}
where \(\lambda_\text{vol} > 0\) is a weighting coefficient. 
This simultaneous optimization creates a powerful synergy: the \(\mathcal{L}_\text{RND}\) term provides a stable, exploratory foundation; \(\mathcal{L}_\text{state}\) distills this exploration into a spatially accurate shortcut; and \(\mathcal{L}_\text{vol}\) ensures this shortcut is also geometrically consistent. 
A pseudocode summarizing the training procedure of OSDS is given by \cref{alg:OSDS-train}.

This principled training scheme produces a high-quality PF ODE flow map. 
At inference time, this allows for both rapid sample generation by evaluating the map, and stable estimation of \(\log Z\) by using the deterministic-flow importance weights. 
Thus, the dual goals of speed and statistical reliability are simultaneously achieved.
The sampling procedure is described in \cref{alg:OSDS-infer}.

\begin{algorithm}[t]
\caption{OSDS Training}
\label{alg:OSDS-train}
\begin{algorithmic}[1]
\REQUIRE densities $\rho$, $\pprior$; horizon $T$, SDE coeffs $f(t),\sigma(t)$; base steps $N_0$ ($d_0=T/N_0$); step-conditioned control $u_\theta$; weight $\lambda_{\text{vol}}$; iterations $I$; batch size $B$.
\ENSURE trained parameters $\theta^\star$.
\STATE Initialize $\theta$
\FOR{$\text{i}=1$ {\bf to} $I$}
    \STATE Draw $\{x_0^{(b)}\}_{b=1}^B \sim \pprior$ 
    \STATE Simulate paths $\{x_{0:N}^{(b)}\}_{b=1}^B$ and compute $\mathcal{L}_\text{RND}$ via \cref{alg:fb-rnd-bound}
    \STATE Sample anchors $(x_t^{(b)},t)$ from paths and steps $\{d^{(b)}\}_{b=1}^B$
    \STATE Compute $\mathcal{L}_\text{state}$ and $\mathcal{L}_\text{vol}$ via \cref{alg:distill}.
    \STATE $\mathcal{L}_{\text{OSDS}} \gets \mathcal{L}_{\text{RND}} + \mathcal{L}_{\text{state}} + \lambda_{\text{vol}}\mathcal{L}_{\text{vol}}$
    \STATE Update $\theta$ by SGD/Adam.
\ENDFOR
\end{algorithmic}
\end{algorithm}

\begin{algorithm}[t]
\caption{OSDS Inference: single/few–step sampling and $\log Z$}
\label{alg:OSDS-infer}
\begin{algorithmic}[1]
\REQUIRE trained $u_\theta$; densities $\rho$, $\pprior$; horizon $T$; test steps $N_{\text{test}}\in\{1,2,4,\ldots\}$; batch size $M$; solver~$\Psi$.
\ENSURE samples $\{y^{(i)}\}$; weights $\{w^{(i)}\}$; $\widehat Z$; $\text{ELBO}$.
\STATE Set step size $d \gets T/N_{\text{test}}$
\STATE Set PF ODE $b_\theta \gets (\tfrac12\sigma u_\theta - f)$
\FOR{$i=1$ {\bf to} $M$} 
    \STATE $x\sim \pprior$;\quad $t \gets 0$;\quad $x_0 \gets x$;\quad $\text{logdet}\gets0$
    \FOR{$j=1$ {\bf to} $N_{\text{test}}$}
        \STATE Integrate PF ODE $x_t \gets \Psi_\theta(x_t,t,d)$
        \STATE Accumulate $\ell \gets \int_t^{t+d} \nabla\cdot b_\theta(x_t,\tau,d)\deriv\tau$ 
        \STATE $\text{logdet}\gets\text{logdet}+\ell$;\quad $t\gets t{+}d$.
    \ENDFOR
    \STATE $y^{(i)}\gets x_t$
    \STATE $\log w^{(i)} \gets \log\rho(y^{(i)}) + \text{logdet} - \log\pprior(x_0)$
    \STATE $w^{(i)} \gets e^{\log w^{(i)}}$
\ENDFOR
\STATE $\widehat Z \gets \frac{1}{M}\sum_i w^{(i)}$;\quad $\text{ELBO}_{\text{det}} \gets \frac{1}{M}\sum_i \log w^{(i)}$.
\end{algorithmic}
\end{algorithm}

\section{EXPERIMENTS}
\subsection{Setup}\label{sec:exp-setup}
We evaluate the performance of OSDS on a diverse set of benchmarks. 
For synthetic targets, we consider the Funnel distribution, the 32‑mode Many‑Well distribution, and a Gaussian mixture model with 40 random modes. 
For real‑world Bayesian inference, we include six standard benchmarks: \textsc{credit}, \textsc{seeds}, \textsc{cancer}, \textsc{Brownian}, \textsc{ionosphere}, and \textsc{sonar}. 
The code is available at \href{https://github.com/PascalJD/one-step-diffusion-samplers}{this GitHub repository}.

\paragraph{Importance weights.}
All evaluation criteria are functions of unnormalized importance weights.
For each sampler, step budget, and run we draw $m$ trajectories and form a set of weights $\{w^{(i)}\}_{i=1}^m$ that compare the target $\rho$ to the density induced by the sampler. 
The weight $w^{(i)}$ is either the discrete forward–backward (RND) path weight (\cref{prop:disc-fb}) or the deterministic‑flow (DF) change‑of‑variables weight (\cref{prop:det-flow-weight}).

\paragraph{Sinkhorn distance.}
To quantify sample quality when we have access to target samples, we report the entropy‑regularized 2‑Wasserstein (Sinkhorn) distance~\citep{cuturi2013sinkhorn} between the empirical distribution of model samples and target samples.
Lower values indicate closer agreement between the generated and target samples.

\paragraph{ELBO (reverse evidence lower bound).}
Following \citet{blessing2024elbos}, we treat the sampler as an importance sampler with weights $w^{(i)}$ and define the evidence lower bound
\[
\text{ELBO} := \E[\log w] \approx \frac{1}{m}\sum_{i=1}^m \log w^{(i)}.
\]
This quantity lower‑bounds $\log Z$ and is high when the importance weights are well calibrated and low-variance.

\paragraph{EUBO (forward evidence upper bound).}
When we can draw samples from the normalized target $\ptarget$, we can also report the forward evidence upper bound (EUBO) \citep{blessing2024elbos}. 
Let $q$ denote the (unknown) model density and $w_\text{fwd}(x)=\rho(x)/q(x)$ the corresponding forward importance weight.
Then
\begin{equation*}
    \begin{aligned}
        \text{EUBO} &:= \E_{x\sim \ptarget}\big[\log w_\text{fwd}(x)\big] \\
        &= \log Z + \KL\big(\ptarget \| q\big) \ge \log Z.
    \end{aligned}
\end{equation*}
Smaller EUBO therefore indicates better mode coverage of the target.
In practice we approximate this expectation by Monte Carlo using the same sampler‑induced weights.

\paragraph{Effective sample size (ESS).}
To assess how concentrated the weights are, we compute the normalized (reverse) effective sample size
\[
\mathrm{ESS} := \frac{\Big(\sum_{i=1}^m w^{(i)}\Big)^2}{m \sum_{i=1}^m \big(w^{(i)}\big)^2}\in (0,1].
\]
Values near $1$ indicate nearly uniform weights (low variance), while values near $0$ signal weight degeneracy.

\paragraph{Partition function error $\boldsymbol{|\Delta \log Z|}$.}
Whenever the true partition function $Z$ is known (for synthetic targets), we form the standard importance‑weighted estimator
\[
\widehat Z = \frac{1}{m}\sum_{i=1}^m w^{(i)},\qquad\widehat{\log Z} = \log \widehat Z,
\]
and report the absolute error
\[
\bigl|\Delta \log Z\bigr| = \bigl|\log Z - \widehat{\log Z}\bigr|.
\]

\paragraph{Baselines and protocols.}
We compare against SMC \citep{delmoral2006smc}, CRAFT \citep{matthews2022craft},
CMCD \citep{vargas2024cmcd}, PIS \citep{zhang2022pis}, and DDS \citep{vargas2023dds}.

For fairness, PIS, DDS, DIS, and our OSDS all share the same PIS‑GRAD backbone \citep{zhang2022pis}, with OSDS further including a small step‑size embedding. 
Each method is evaluated over 20 independent runs, with 2{,}000 samples per run; we report mean~$\pm$~standard deviation across runs. 
All samplers use the same right‑endpoint Euler-Maruyama discretization. 
Further implementation details are deferred to \cref{sec:model-opt}.

\subsection{Sample Quality}
\paragraph{Sinkhorn distance.}
\Cref{tab:synthetic-sinkhorn} summarizes Sinkhorn distance between model samples and the target on two synthetic benchmarks. 
Even with a single step, OSDS attains competitive distances: on \textsc{Funnel} it is within $\sim$0.8\% of PIS  while using $1/128$ of the function evaluations; on \textsc{MW} OSDS is close in absolute terms to the best diffusion samplers (0.39 vs.\ 0.14 for PIS/DDS). 
These results indicate that our PF flow map preserves much of the sample quality typically associated with iterative diffusion/bridge samplers, despite using dramatically fewer steps.

\begin{table}[t]
\centering
\setlength{\tabcolsep}{3pt}
\renewcommand{\arraystretch}{1.08}
\caption{Sinkhorn distance on synthetic targets. OSDS attains competitive sample quality in a single step.}
\label{tab:synthetic-sinkhorn}
\begin{tabular}{llcc}
\toprule
Sinkhorn \((\downarrow)\)& NFE &
\makecell[c]{Funnel (10D)} &
\makecell[c]{MW (5D)} \\
\midrule
SMC & 128 & 149.35\std{4.73}  & 20.71\std{5.33} \\
CRAFT & 128 & 133.42\std{1.04}  & 11.47\std{0.90} \\
CMCD & 128 & 124.89\std{8.95}  & 0.57\std{0.05} \\
PIS   & 128 & 159.13\std{0.34}  & 0.14\std{0.00} \\
DDS   & 128 & 147.72\std{0.90}  & 0.14\std{0.00} \\
\addlinespace[3pt]
OSDS  (ours)&   1 & 160.36 \std{0.64} & 0.39 \std{0.00} \\
\bottomrule
\end{tabular}
\end{table}

\begin{table*}
\centering
\caption{ELBO from importance weights on six Bayesian benchmarks. DF denotes our proposed deterministic-flow change-of-variables weights (PF ODE) while RND denotes the discrete stochastic forward-backward path weights.}
\label{tab:elbos}
\setlength{\tabcolsep}{2pt}
\renewcommand{\arraystretch}{0.95}
\small
\begin{tabular}{@{}l l l c c c c c c@{}}
\toprule
ELBO \((\uparrow)\)& Weight & NFE & Credit (25D) & Seeds (26D) & Cancer (31D) & Brown. (32D) & Iono.(35D) & Sonar (61D)\\
\midrule
OSDS (ours)& DF & 1   & -777.22\std{8.64}                     &   -89.97\std{1.00}     &   -2869\std{90}   & -11.37\std{0.12}                 &  -232.70\std{1.00}  &  -304.40\std{3.00} \\
OSDS (ours) & RND & 1   & -20758.59\std{540.69}                     & -9892\std{2.7e+03} & -10640\std{3e+02} & -278.51\std{2.91}                  & -1302\std{8.9}  & -2010\std{18.00} \\
\addlinespace[3pt]
OSDS (ours) & DF & 128 & \textbf{-501.74}\std{0.53}  & \textbf{-44.35}\std{0.56} & 8.66\std{0.19} & -11.58\std{0.13} & \textbf{-86.68}\std{0.72} & \textbf{-50.58}\std{0.56} \\
OSDS (ours)  & RND  & 128 & -519.73\std{0.19}  & -79.74\std{0.10} & 2.90\std{0.16} & -32.62\std{0.14} & -124.83\std{0.24} & -139.70\std{0.28} \\
\addlinespace[3pt]
PIS & RND& 128 & -846.57\std{2.42}  & -88.92\std{2.05} & \textbf{39.54}\std{5.30} & NA & -125.03\std{0.69} & -142.868\std{3.29} \\
DDS & RND & 128 & -514.74\std{1.22} & -75.21\std{0.21} & 20.00\std{0.690} & \textbf{0.51}\std{0.23} & -114.19\std{0.11} & -121.22\std{5.99} \\
\bottomrule
\end{tabular}
\end{table*}

\begin{table*}[t]
\centering 
\caption{Comparison of our proposed deterministic-flow (DF) and discrete forward–backward (RND) importance weights on the 40-mode Gaussian mixture model across different step budgets (NFE).}
\label{tab:df-gmm}
\begin{tabular}{rcccccccc}
\toprule
 & \multicolumn{2}{c}{ELBO \((\uparrow)\)} 
 & \multicolumn{2}{c}{EUBO \((\downarrow)\)} 
 & \multicolumn{2}{c}{ESS \((\uparrow)\)} 
 & \multicolumn{2}{c}{$|\Delta \log Z|$ \((\downarrow)\)} \\
NFE & DF & RND & DF & RND & DF & RND & DF & RND \\
\midrule
1   & -9.69\std{0.41} &  -11532.86\std{362.64} & 1.66\std{0.02} & 17.36\std{0.40} & 0.05\std{0.01} & 0.00\std{0.00} & 0.31\std{0.36} & 4.08\std{0.57} \\
2   & -3.36\std{0.20} &  -2646.03\std{84.56}  & 1.78\std{0.02} &  5.608\std{0.07} & 0.13\std{0.03} & 0.00\std{0.00} & 0.70\std{0.07} & 2.57\std{0.50}  \\
4   & -3.45\std{0.27} & -66.89\std{3.22} & 2.08\std{0.02} & 3.28\std{0.04} & 0.41\std{0.03} & 0.02\std{0.01} & 0.97\std{0.03} & 1.47\std{0.38} \\
8   & 1.39\std{0.16} & -19.72\std{0.59} & 2.00\std{0.02} & 3.00\std{0.05} & 0.36\std{0.08} & 0.03\std{0.01} & 0.75\std{0.04} & 1.00\std{0.36} \\
16  & 2.31\std{0.05} & -5.43\std{0.13} & 2.05\std{0.03} & 2.18\std{0.04} & 0.61\std{0.03} & 0.06\std{0.02} & 0.99\std{0.02} & 0.32\std{0.22} \\
32  & 2.64\std{0.04} & -2.09\std{0.06} & 2.15\std{0.02} & 1.45\std{0.03} & 0.65\std{0.04} & 0.10\std{0.04} & 0.78\std{0.02} & 0.11\std{0.09} \\
64  & 3.10\std{0.04} & -1.98\std{0.06}  & 2.91\std{0.02} &  1.83\std{0.06} & 0.64\std{0.01} & 0.17\std{0.05} & 0.77\std{0.05} & 0.05\std{0.05} \\
128 & 3.59\std{0.29} & -2.92\std{0.14}  & 2.07\std{0.02} &  2.94\std{0.05} & 0.66\std{0.01} & 0.23\std{0.06} & 0.73\std{0.02} & 0.01\std{0.09} \\
\bottomrule
\end{tabular}
\end{table*}

\begin{figure}[t]
\centering
\includegraphics[width=0.48\textwidth]{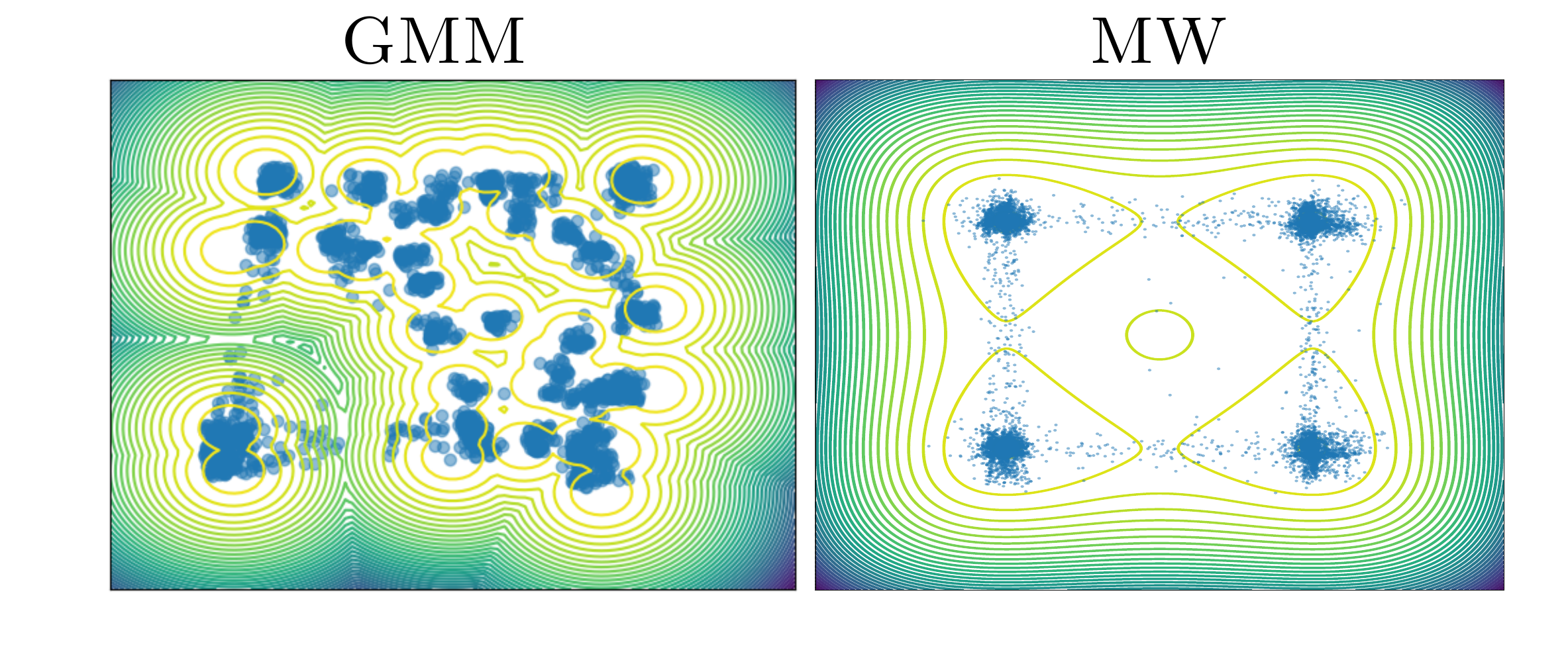}
\caption{Single‑step OSDS samples (blue) overlaid on target energy contours. Samples populate distinct high‑density regions, illustrating broad mode coverage in one step.}
\label{fig:mode-coverage}
\end{figure}

\paragraph{Mode coverage.}
\Cref{fig:mode-coverage} visualizes OSDS’s single-step samples overlaid on energy contours for a Gaussian mixture model (GMM) with 40 random modes in two dimensions and a five-dimensional 32-mode Many-Well (MW). 
On \textsc{GMM}, mass is spread across the mixture components, while on \textsc{MW} the samples occupy all high‑probability wells. 
These qualitative results mirror the Sinkhorn findings: OSDS maintains coverage of modes in one step, aligning samples with the high‑density regions of the target energy. 

\paragraph{Trading compute for sample quality.}
While our main comparisons focus on the single-step regime, OSDS also exposes a knob to trade additional NFEs for tighter inference.
On the 40‑mode GMM in \Cref{tab:df-gmm}, increasing the PF‑ODE step budget improves the deterministic‑flow (DF) weights and recovers the accuracy of the RND estimator in the small step regime. 

\paragraph{Amortized inference cost.}
Although OSDS introduces a small distillation overhead during training, this cost is amortized at inference. 
In Appendix~\ref{sec:cost-analysis}, we derive an explicit NFE accounting that compares a baseline $N$-step diffusion sampler with OSDS. 
Our analysis shows that for moderate or large sampling workloads, the one-time distillation cost is quickly recovered, yielding substantial end-to-end computational savings.

\subsection{Deterministic-Flow Importance Weights}\label{sec:df-weights}
We now evaluate the proposed deterministic-flow (DF) importance weights and compare them against the standard discrete forward-backward path weights (RND). 
We consider six Bayesian benchmarks from \citep{blessing2024elbos} and the strongly multimodal 40‑mode GMM.

\paragraph{Bayesian benchmarks.}
Table~\ref{tab:elbos} reports ELBOs obtained from DF and RND weights for OSDS at two discretization budgets: a single PF-ODE step and a fine discretization, alongside PIS and DDS baselines.

In the \emph{single-step} regime, DF remains numerically stable on all six datasets, yielding finite, reasonably tight ELBOs. 
In contrast, the RND estimator collapses: ELBO values are several orders of magnitude lower, reflecting the forward-backward kernel mismatch at coarse resolution. 
This matches our analysis in \cref{sec:em-example}: when the discretization is too coarse, the surrogate backward kernel is no longer time-adjoint to the forward kernel, and path-space likelihood ratios become ill-conditioned, whereas DF sidesteps the backward kernel entirely.

In the \emph{multi-step} regime, DF continues to improve and often surpasses diffusion baselines. 
OSDS (DF, 128) achieves the best ELBO on \textsc{credit}, \textsc{seeds}, \textsc{ionosphere}, and \textsc{sonar}, and remains competitive on \textsc{cancer} and \textsc{Brownian}. 
At the same step budget, RND consistently yields lower ELBOs than DF, indicating higher-variance weights even when the discretization is fine. 
This suggests that enforcing geometric consistency via the PF ODE and volume tracking leads to better-conditioned importance weights and tighter evidence estimates.

\paragraph{Multimodal 40‑mode GMM.}\label{sec:gmm40}
To further stress-test the importance weights in a strongly multimodal setting, we consider the 40‑mode Gaussian mixture model and compare DF and RND across a range of step budgets. 
Table~\ref{tab:df-gmm} reports ELBO, EUBO, ESS, and $|\Delta \log Z|$ as a function of NFE for both estimators.

In the \emph{few-step} regime (NFE $=1,2,4,8$), the RND estimator exhibits the breakdown predicted by our theory. 
ELBO values are extremely negative, the ESS is essentially zero, and $|\Delta \log Z|$ is large. 
By contrast, DF remains well behaved under exactly the same discretizations: ELBOs are moderate and improve rapidly with NFE, EUBO stays close to the ground-truth scale, ESS is consistently non-degenerate, and $|\Delta \log Z|$ is already below~1 for all NFE in this regime. 
This confirms that DF provides stable and accurate importance weights precisely where path-space RND weights become unreliable.

As the number of steps increases, the RND estimator gradually recovers. 
For NFE $\geq 16$, its $|\Delta \log Z|$ becomes very small, reflecting the asymptotic correctness of the discrete forward-backward likelihood ratio when the integrator is sufficiently fine. 
However, even in this multi-step regime, RND weights remain high-variance: ESS is consistently lower than for DF, and the resulting ELBOs are also lower. 
DF, on the other hand, achieves strong ELBOs and ESS while maintaining $|\Delta \log Z|$ below~1 over the entire NFE range.
Overall, these results show that DF importance weights are robust in the one/few-step regime, where discrete forward-backward estimators can catastrophically fail, and remain competitive in the multi-step regime.

\subsection{Ablation on the volume-consistency weight}\label{sec:ablation-volume}
To assess the influence of the volume-consistency loss $\mathcal{L}_\text{vol}$, we ablate its weight $\lambda_\text{vol}$ on one synthetic target (40‑mode GMM) and one Bayesian benchmark (\textsc{credit}). 
Table~\ref{tab:vol-ablation} reports $\lambda_\text{vol}=0$ (no volume consistency) and $\lambda_\text{vol}=0.25$. 
In all cases, adding a small amount of volume regularization improves the ELBO, with larger gains in the challenging one‑step regime.

\begin{table}[t]
\centering
\caption{Ablation on the volume-consistency weight $\lambda_\text{vol}$ using deterministic-flow (DF) ELBO on the 40-mode GMM and the \textsc{credit} Bayesian benchmark.}
\label{tab:vol-ablation}
\setlength{\tabcolsep}{6pt}
\renewcommand{\arraystretch}{1.05}
\begin{tabular}{lccc}
\toprule
Task & NFE & $\lambda_\text{vol}$ & ELBO $(\uparrow)$ \\
\midrule
GMM           & 1   & 0    & $-10.74$\std{0.41} \\
GMM           & 1   & 0.25 & $\mathbf{-9.69}$\std{0.30} \\
GMM           & 128 & 0    & $2.19$\std{0.06} \\
GMM           & 128 & 0.25 & $\mathbf{3.59}$\std{0.29} \\
\addlinespace[2pt]
\textsc{credit} & 1   & 0    & $-1555.76$\std{15.81} \\
\textsc{credit} & 1   & 0.25 & $\mathbf{-777.22}$\std{8.64} \\
\textsc{credit} & 128 & 0    & $-509.32$\std{0.49} \\
\textsc{credit} & 128 & 0.25 & $\mathbf{-501.74}$\std{0.53} \\
\bottomrule
\end{tabular}
\end{table}

\section*{CONCLUSION}
We introduced Self-Distilled One-Step Diffusion Samplers (OSDS), a framework that reconciles fast inference with statistically sound evidence estimation for sampling from unnormalized densities. 
Our analysis identified a root cause of ELBO collapse in the few-step regime, the time-asymmetry of standard discrete integrators, which breaks the forward-backward likelihood ratio. 
OSDS addresses this by distilling many small probability-flow steps into a geometrically consistent shortcut map, enforced by state and volume consistency losses, and by estimating evidence via a deterministic change-of-variables weight that bypasses the fragile backward kernel. 
Empirically, OSDS delivers stable single-step ELBOs and strong sample quality.

\section*{ACKNOWLEDGEMENTS}
This research is supported in part by NSF IIS-2508145 and Amazon Research Award.

\bibliography{references}

\appendix
\thispagestyle{empty}

\onecolumn
\aistatstitle{Supplementary Materials}

\section{Notation and Assumptions}\label{sec:notation}
Throughout we fix a filtered probability space \((\Omega,\mathcal F,(\mathcal F_t)_{t\in[0,T]},\mathbb P)\) supporting a $d$‑dimensional standard Wiener process \(\rvw=(\rvw_t)_{t\in[0,T]}\).
All stochastic integrals are in the Itô sense unless stated otherwise.

We denote by \(\rvx = (\rvx_t)_{t\in[0,T]}\in C([0,T],\R^D)\) the forward diffusion process governed by the SDE
\cref{eq:generative-sde}.  
The law of the entire path $\rvx_{0:T}$ is written $\sP^{\pi,u}$ when the initial density is $\pi$ and the control (drift correction) is $u$.
Analogously, $\backvec{\sP}^{\tau,v}$ denotes the law of the reverse‑time SDE \cref{eq:noising-sde} with terminal density
$\tau$ and control $v$.

The unnormalized target density is \(\rho:\R^D\to(0,\infty)\) with unknown partition function \(Z=\int\rho(x)\deriv x\).
The normalized target is \(\ptarget=\rho/Z\).
The prior (reference) density is \(\pprior\).
Unless otherwise noted, expectations \(\E[\cdot]\) are taken with respect to the forward path measure \(\sP^{\pi,u}\).

$\mathcal U \subset C(\R^D\times[0,T],\R^D)$ is the set of admissible controls. 
A subscript $\theta$ (e.g.\ $u_\theta$) indicates a neural‑network parameter vector \(\theta\in\R^p\).

For numerical schemes we use a mesh \(0=t_0<t_1<\dots<t_N=T\) and write \(\Delta t_n=t_{n+1}-t_n\). 
The shorthand \(\rvx_{0:N}:=(\rvx_{t_0},\dots,\rvx_{t_N})\) denotes the sampled discrete path; \(p_{n+1|n}\) refer to the corresponding Markov transition densities.

We assume that $x\mapsto f(x,t)$ and $x\mapsto u(x,t)$ are globally Lipschitz with at most linear growth, uniformly in $t\in[0,T]$.
We assume $\sigma(t)$ is continuous and bounded away from $0$ and $\infty$ on $[0,T]$.
These standard conditions guarantee strong existence and uniqueness for \cref{eq:generative-sde,eq:noising-sde} \cite[Theorem 5.2.1]{Oksendal2003}, absolute continuity of the induced path measures, and finiteness of the Radon-Nikodym derivatives we use.

We use the dot notation $\dot{\rvx}_t := \frac{\deriv}{\deriv t}\rvx_t$ for time derivatives.
Given a (learned) vector field $b_\theta:\R^D\times[0,T]\to\R^D$, the probability flow ODE (PF ODE) associated with the generative SDE is
\[
\dot{\rvx}_t=b_\theta(\rvx_t,t),\qquad \rvx_0=x_0.
\]
Its flow map $\phi_t:\R^D\to\R^D$ satisfies $\phi_0(x)=x$ and $\frac{\deriv}{\deriv t}\phi_t(x)=b_\theta(\phi_t(x),t)$.
For a transport horizon $d>0$ we write $T:=\phi_d$ for the single-step deterministic map.
We use $\nabla T(x)$ for the Jacobian, $J_T(x):=\nabla T(x)$, and $\log \det$ for the log-determinant.

Along a PF ODE trajectory we define the scalar accumulator
$\ell_t$ by the auxiliary ODE
\[
\dot{\ell}_t=\nabla\cdot b_\theta(\rvx_t,t),\qquad \ell_0=0,
\]
so that the log-Jacobian of $T=\phi_d$ along the trajectory starting at $x_0$ is $\log|\det \nabla T(x_0)|=\ell_d$ \citep{chen2018ctnf}.

\section{Radon-Nikodym Derivative}
\subsection{Forward-backward Radon-Nikodym derivative}\label{sec:rnd}
Consider forward and reverse-time stochastic processes on \(\sR^D\) over a time interval \([0, T]\), each described by the following SDEs:
\begin{align}
    \deriv \rvx_t &= (f + \sigma u)(\rvx_t, t)\deriv t + \sigma(t){\deriv} \rvw_t, \quad \rvx_0 \sim \pi, \label{eq:forward-sde}\\
    \deriv \rvx_t &= (-\backvec f + \backvec\sigma \backvec v)(\rvx_t, t)\deriv t + \backvec\sigma(t)\deriv  \rvw_t, \quad \rvx_0 \sim \tau. \label{eq:reverse-sde}
\end{align}
\paragraph{Proposition~\ref{prop:rnd} (Forward-backward Radon-Nikodym derivative), restated.}
For \(u, v \in \mathcal{U} \subset C(\sR^D \times [0,T], \sR^D)\), \(\mu \in C(\sR^D \times [0,T], \sR^D)\) and \(\sigma \in C([0,T], \sR)\). 
The likelihood ratio between the path measures \(\sP^{\pi, u}\) and \(\backvec \sP^{\tau, v}\) of the processes \cref{eq:forward-sde} and \cref{eq:reverse-sde}, respectively, is given by
\begin{equation}\label{eq:generic-rnd}
    \begin{aligned}
        \log\frac{\deriv {\sP^{\pi, u}} }{\deriv \backvec \sP^{\tau, v}}(\rvx) &= \frac{1}{2}\int_0^T\left(\|\sigma f\|^2 - \|\sigma f + u\|^2\right)(\rvx_t, t)\deriv t + \int_0^T\left(\frac{u}{\sigma}\right)(\rvx_t, t)\cdot\deriv \rvx_t \\
        &+ \frac{1}{2}\int_0^T\left(\|\sigma f + v\|^2 - \|\sigma f\|^2\right)(\rvx_t, t)\deriv t - \int_0^T \left(\frac{v}{\sigma}\right)(\rvx_t, t)\cdot\backvec{\deriv}\rvx_t\\
        &+ \log\frac{\pi(\rvx_0)}{\tau(\rvx_T)}.
    \end{aligned}
\end{equation}

\begin{proof}
    The result follows the structure of \cite[proof of Proposition 2.2]{vargas2024cmcd}. We introduce a reference process with control \(w\) and define the corresponding path measures \({\sP}^{\pi, w}\), \(\backvec{\sP}^{\tau, w}\).
    The Radon-Nikodym derivative can be decomposed as the forward and backward Radon-Nikodym derivatives:
    \begin{equation}\label{eq:rnd-decomposition}
         \log \frac{\deriv {\sP}^{\pi, u}}{\deriv \backvec{\sP}^{\tau, v}}(\rvx) = \log \frac{\deriv {\sP}^{\pi, u}}{\deriv {\sP}^{\pi, w}}(\rvx) + \log\frac{\deriv \backvec{\sP}^{\tau, w}}{\deriv \backvec{\sP}^{\tau, v}}(\rvx) + \log\frac{\pi(\rvx_0)}{\tau(\rvx_T)}.
    \end{equation}
    The result then follows by applying Girsanov theorem \cite[Section A.1]{nusken2021solving} forward and backward in time.
    The forward term can be expressed as
    \begin{equation*}
        \log \frac{\deriv {\sP}^{\pi, u}}{\deriv {\sP}^{\pi, w}}(\rvx) = \frac{1}{2}\int_0^T\left(\|\sigma f + w\|^2 - \|\sigma f + u\|^2\right)(\rvx_t, t)\deriv t + \int_0^T\left((u - w)\sigma^{-1}\right)(\rvx_t, t)\cdot\deriv \rvx_t
    \end{equation*}
    Similarly, the backward term can be expressed as:
    \begin{align*}
        \log\frac{\deriv \backvec{\sP}^{\tau, w}}{\deriv \backvec{\sP}^{\tau, v}}(\rvx) 
        &= \log\frac{\deriv {\sP}^{\tau, w}}{\deriv {\sP}^{\tau, v}}(\rvx) \\
        &- \int_0^T \left((w-v)\sigma^{-1}\right)(\rvx_t, t)\cdot {\deriv}\rvx_t + \int_0^T \left((w-v)\sigma^{-1}\right)(\rvx_t, t)\cdot\backvec{\deriv}\rvx_t\\
        &= \frac{1}{2}\int_0^T\left(\|\sigma f+ v\|^2 - \|\sigma f + w\|^2\right)(\rvx_t, t)\deriv t + \int_0^T \left((w-v)\sigma^{-1}\right)(\rvx_t, t)\cdot\backvec{\deriv}\rvx_t
    \end{align*}
    Setting \(w := 0\) and plugging into \cref{eq:rnd-decomposition} yields the Radon-Nikodym derivative \cref{eq:rnd}.
\end{proof}

\subsection{Discrete forward-backward likelihood ratio and a lower bound on \texorpdfstring{$\log Z$}{log Z}}
\label{sec:disc-fb}

We discretize the time interval $[0,T]$ by \(0=t_{0}<t_{1}<\dots<t_{N}=T.\)

\paragraph{Proposition~\ref{prop:disc-fb} (Finite-step forward-backward bound), restated.}
Let \(\sP^{\pi,u}\) be the path measure of the forward SDE~\cref{eq:forward-sde} driven by a (possibly approximate) control \(u\), and let \(\backvec{\sP}^{\tau,v}\) be the reverse‑time SDE~\cref{eq:reverse-sde} with terminal density \(\tau=\rho/Z\).  
Denote by
\[
    p_{n+1|n}(\rvx_{n+1}\mid\rvx_{n}),\qquad p_{n|n+1}(\rvx_{n}\mid\rvx_{n+1})
\]
the one‑step transition densities of any consistent discretization scheme applied to the two SDEs.  
Then for every discrete trajectory \(\rvx_{0:N}:=(\rvx_{0},\dots,\rvx_{N})\) generated forward in time under \(\sP^{\pi,u}\),
\begin{equation}\label{eq:disc-fb-rnd} 
\log\frac{\deriv \sP^{\pi,u}}{\deriv \backvec{\sP}^{\tau,v}}(\rvx_{0:N})
   = \log\frac{\pi(\rvx_{0})}{\tau(\rvx_{N})} + \sum_{n=0}^{N-1} \log\frac{p_{n+1|n}(\rvx_{n+1}\mid\rvx_{n})}{p_{n|n+1}(\rvx_{n}\mid\rvx_{n+1})}.
\end{equation}
Consequently
\begin{equation}\label{eq:disc-fb-bound}
    \mathbb{E}_{\sP^{\pi,u}}\Bigl[\sum_{n=0}^{N-1} \log\frac{\rho(\rvx_{N})p_{n+1|n}(\rvx_{n+1}\mid\rvx_{n})}{\pi(\rvx_{0})p_{n|n+1}(\rvx_{n}\mid\rvx_{n+1})} \Bigr]\le\log Z
\end{equation}
with equality when the exact optimal drifts are used and the discretization is exact.  
The inequality holds for any step‑size sequence and for any approximate control \(u\).

\begin{proof}
Factorizing the discrete path densities, \(\sP^{\pi,u}(\rvx_{0:N}) = \pi(\rvx_{0})\prod_{n=0}^{N-1}p_{n+1|n}(\rvx_{n+1}\mid\rvx_{n})\) and \(\backvec{\sP}^{\tau,v}(\rvx_{0:N}) = \tau(\rvx_{N})\prod_{n=0}^{N-1}p_{n|n+1}(\rvx_{n}\mid\rvx_{n+1}),\) gives~\cref{eq:disc-fb-rnd} immediately.  
Taking the expectation of both sides with respect to
\(\sP^{\pi,u}\) and noting that \(\E_{\sP^{\pi,u}}\bigl[\log\dfrac{ \deriv \sP^{\pi,u}}{\deriv  \backvec{\sP}^{\tau,v}}\bigr] = \mathrm{KL}(\sP^{\pi,u}\Vert\backvec{\sP}^{\tau,v})\ge 0\) establishes~\cref{eq:disc-fb-bound}.  
Because the Kullback–Leibler divergence is non‑negative, the bound is always valid, even if \(u\) (or \(v\)) is sub‑optimal or the numerical scheme is coarse.
\end{proof}

In discrete time, the Radon-Nikodym derivative between the path measures of a forward simulator and its reverse factorizes into a boundary term and a forward-over-backward product of one-step Markov kernels (\cref{eq:disc-fb-rnd}).
Here $p_{n+1|n}$ is the forward one-step kernel of the chosen integrator for the generative SDE, while $p_{n|n+1}$ is the corresponding backward kernel (the conditional of the same discrete chain run backward, i.e., conditioned on the right endpoint). 
Crucially, $p_{n|n+1}$ is not in general equal to “the EM step of the reverse-time SDE’’.
See \cite[Remark~3.2]{blessing2025udb} for the same discrete RND viewpoint in the underdamped setting.

\paragraph{Practical Monte‑Carlo estimator.}
Drawing $M$ independent paths under the forward discretization and substituting them into \cref{eq:disc-fb-bound} yields the unbiased estimator
\begin{equation}\label{eq:logz}
    \widehat{\log Z} := \frac1M\sum_{m=1}^{M} \sum_{n=0}^{N-1} \log\frac{\rho(\rvx^{(m)}_{N})p_{n+1|n}(\rvx^{(m)}_{n+1}\mid\rvx^{(m)}_{n})}{\pi(\rvx^{(m)}_{0})p_{n|n+1}(\rvx^{(m)}_{n}\mid\rvx^{(m)}_{n+1})},
\end{equation}
Drawing $M$ paths yields an unbiased estimator of the ELBO (the LHS of \eqref{eq:disc-fb-bound}). 
Since the ELBO lower-bounds \(\log Z\), this provides a lower bound in expectation.

Because the proof only uses (i) absolute continuity of the discrete path measures and (ii) positivity of KL, it remains valid when the continuous SDE is replaced by any approximate simulator.  
Using a more accurate scheme tightens the bound but can never violate~\cref{eq:disc-fb-bound}.

Setting $N=1$ in Prop.~\ref{prop:disc-fb} gives the weight
\begin{equation}\label{eq:disc-fb-single}
  \log\frac{\deriv \sP^{\pi,u}}{\deriv \backvec{\sP}^{\tau,v}}((\rvx_0, \rvx_N)) = \log\frac{\pi(\rvx_{0})}{\tau(\rvx_{N})} + \log\frac{p_{N|0}(\rvx_{N}\mid \rvx_{0})}{p_{0|N}(\rvx_{0}\mid \rvx_{N})},
\end{equation}
i.e.\ the log‑ratio of two Gaussian kernels plus the prior/target
term.  
Because it is still a likelihood ratio, the expectation remains an ELBO even when the single step is large.

\paragraph{Line–integral discretization.}
Many diffusion papers compute the path log‑weight directly from the Girsanov formula \citep{berner2024dis, richter2024improved},
\begin{equation}\label{eq:rnd-li}
    \log\frac{\deriv\sP^{\pi,u}}{\deriv \backvec{\sP}^{\tau,v}} = \int_0^T (u + v) \cdot \Big(u + \frac{v - u}{2} + \nabla \cdot (\sigma v - \mu)\Big)(\rvx_t, t)\deriv t + \int_0^T \big(u + v\big)(\rvx_t, t) \cdot \deriv \rvw_t +  \log \frac{\pi(\rvx_0)}{\tau(\rvx_T)}.
\end{equation}
Discretized as an approximate Riemann sum, this divergence-based Radon-Nikodym derivative does not guarantee a lower-bound for the log-normalization constant \citep{blessing2025udb}. 
When sampling using $N=1$ steps, the inner running cost of the line-integral discretization vanishes and the estimation collapses to the boundary ratios.

\subsection{Why EM breaks at one step}\label{sec:em-example}

\paragraph{EM is popular, but fails at coarse discretization.}
With many small steps, Euler-Maruyama (EM) discretization produces nearly time-symmetric pairs, so each log-ratio term is close to zero and the importance weights are well-behaved.

Write $\Delta t_n=t_{n+1}-t_n$.  
For the forward SDE \cref{eq:generative-sde}, EM gives the forward kernel
\begin{equation}\label{eq:em-forward}
p_{n+1|n}(x'|x)=\mathcal N\Big(x'; x+\big(f(t_n)x+\sigma(t_n)u(x,t_n)\big)\Delta t_n,\sigma(t_n)^2\Delta t_nI\Big).
\end{equation}
Applied to the reverse-time SDE \cref{eq:noising-sde}, the right-point EM update gives
\begin{equation}\label{eq:em-backward}
\widetilde p_{n|n+1}(x|x')=\mathcal N\Big(x;x'-\backvec f(t_{n+1})x'\Delta t_n,\backvec\sigma(t_{n+1})^2\Delta t_nI\Big).
\end{equation}
Both kernels are consistent as \(\max_n\Delta t_n\to 0\). 
At coarse resolution, however, they are highly time-asymmetric, since \cref{eq:em-forward} is calibrated at \((x,t_n)\) and \cref{eq:em-backward} at \((x',t_{n+1})\). 

When \(\Delta t_n\) is large, \(\widetilde p_{n|n+1}\) in \cref{eq:em-backward} is not the time-adjoint of \cref{eq:em-forward}. 
\begin{definition}[Time-adjoint.]
    Given a forward kernel \(p_{n+1|n}\) and a marginal \(\pi_n\), the time-reversed kernel is
\begin{equation*}\label{eq:time-adjoint}
    p^{\star}_{n|n+1}(x|x')=\frac{\pi_n(x)p_{n+1|n}(x'|x)}{\int \pi_n(z)p_{n+1|n}(x'|z)\deriv z}.
\end{equation*}
We call \(p_{n+1|n}\) and \(p_{n|n+1}\) \emph{time-adjoint} if \(p_{n|n+1}=p^{\star}_{n|n+1}\) for the \(\pi_n\) transported by \(p_{n+1|n}\). 
\end{definition}
If \(p_{n|n+1}=p^{\star}_{n|n+1}\), then the discrete forward–backward (FB) Radon-Nikodym ratio \cref{eq:disc-fb-rnd-main} is an exact likelihood ratio for the discrete path measures and its expectation gives a valid ELBO. 

Replacing \(p^{\star}_{n|n+1}\) by \(\widetilde p_{n|n+1}\) introduces a cross-entropy gap at each step:
\begin{equation*}\label{eq:entropy-gap}
        \E\big[\log p_{n+1|n}(\rvx_{n+1}| \rvx_n)-\log \widetilde p_{n|n+1}(\rvx_n| \rvx_{n+1})\big] = -\E_{\rvx_{n+1}}\Big[\mathrm{KL}\big(p^{\star}_{n|n+1}(\cdot|\rvx_{n+1})\|\widetilde p_{n|n+1}(\cdot|\rvx_{n+1})\big)\Big]\le 0.
\end{equation*}
At \(N=1\) this single KL term can be very large because the forward EM mean and variance scale with \(\beta(t_1)\) at the right endpoint and with the control \(u\), whereas a surrogate backward that ignores these choices concentrates elsewhere. 
The ELBO and estimation of \(\log Z\) become overly conservative, even when the samples remain visually plausible.

\paragraph{A 1D example.}
Consider the VP SDE family with constant \(\beta>0\) and \(\sigma(t)=\sigma_0\sqrt{\beta}\), and take \(u_\theta\equiv 0\) for clarity.
We discretize \([0,1]\) with a single step \(\Delta t=1\).

The forward SDE uses drift \(+\tfrac12\beta x\), so one EM step yields the linear-Gaussian update
\[
x_1 = Ax_0 + \eta,\qquad A:= 1 + \tfrac12\beta,\quad \eta \sim \mathcal N(0,Q),\quad Q := \beta\sigma_0^2.
\]

Suppose \(x_0\sim\mathcal N(0,\Sigma_0)\) with \(\Sigma_0=\sigma_0^2\) (the usual VP prior).

Then the true backward Markov kernel of this discrete forward chain is the Kalman posterior
\[
p_{\text{true}}(x_0 | x_1) = \mathcal N\bigl(Kx_1,\Sigma_{\text{post}}\bigr),\qquad
K := \frac{\Sigma_0 A}{A^2 \Sigma_0 + Q} = \frac{1 + \tfrac12\beta}{(1 + \tfrac12\beta)^2 + \beta},
\]
\[
\Sigma_{\text{post}} = \Sigma_0 - \frac{\Sigma_0^2 A^2}{A^2\Sigma_0 + Q}
= \sigma_0^2 \cdot \frac{\beta}{(1 + \tfrac12\beta)^2 + \beta}.
\]

By contrast, the “EM backward” kernel one obtains by applying a right‑endpoint EM step to the reverse direction is
\[
p_{\text{em-bwd}}(x_0 | x_1) = \mathcal N\bigl((1-\tfrac12\beta)\,x_1,\;Q\bigr).
\]

For a typical value \(\beta=10\) and \(\sigma_0^2=1\) we get
\[
A=6,\quad Q=10,\quad K=\frac{6}{36+10} \approx 0.1304,\quad \Sigma_{\text{post}} = \frac{10}{46} \approx 0.217.
\]

The two conditionals differ by more than an order of magnitude in both mean and variance:
\[
\text{mean: } Kx_1 \approx 0.13x_1 \quad\text{vs}\quad (1-\tfrac12\beta)x_1 = -4x_1,
\]
\[
\text{variance: } \Sigma_{\text{post}} \approx 0.217 \quad\text{vs}\quad Q=10.
\]

Hence for fixed \(x_1\), \(p_{\text{true}}(x_0|x_1)\) is a tight Gaussian near \(0.13x_1\), while \(p_{\text{em-bwd}}(x_0|x_1)\) is an extremely broad Gaussian centered at \(-4\,x_1\).

The per‑step mismatch can be quantified exactly:
\[
\E_{x_0\mid x_1}[\log p_{\text{true}}(x_0\mid x_1)-\log p_{\text{EM-bwd}}(x_0\mid x_1)]
= \mathrm{KL}(p_{\text{true}}\|\;p_{\text{EM-bwd}})\;\ge 0
\]

The conditional KL: \( \E \big[\mathrm{KL}(p^\star(\cdot\mid x_1)\|p_{\text{EM-bwd}}(\cdot\mid x_1))\big] \approx 40 \text{ nats}\). 
By \eqref{eq:entropy-gap-main}, this subtracts roughly 40 nats from the per-step FB-RND contribution when we use $p_{\text{EM-bwd}}$ instead of the time-adjoint $p^\star$.

Therefore, replacing the true discrete backward kernel by “EM backward” breaks the forward-backward identity at one step and catastrophically degrades the ELBO and \(\log Z\) estimator.

In discrete time, the FB Radon-Nikodym ratio requires the backward Markov kernel of the same discrete chain used forward.
At \(N=1\), EM’s right‑point forward step and its right‑point “backward” step are not a time‑adjoint pair, so the per‑step log‑ratio is dominated by a large cross‑entropy term between the true and surrogate conditionals.
This is negligible at small \(\Delta t\) but becomes dominant at one or a few steps.

\section{Deterministic-flow importance weights for single-step estimation of \texorpdfstring{$Z$}{Z}}
\label{sec:det-flow-weight}

\subsection{Deterministic \texorpdfstring{$\log Z$}{log Z} estimate}\label{sec:det-logz}
We show how to compute an importance weight from a deterministic transport induced by the probability-flow (PF) ODE. 
This enables a single-step estimator of the partition function \(Z=\int_{\R^D}\rho(y)\deriv y\) using samples from the prior \(\pprior\).

\textbf{\Cref{prop:det-flow-weight} (Deterministic-flow IS weight), restated}
Let \(q:=T_{\#}\pprior\) be the push-forward of the prior by the flow map \(T=\phi_d\). Then for any $x_0\in\R^D$,
\begin{equation}
\label{eq:det-flow-weight}
w(x_0) = \frac{\rho \big(T(x_0)\big)}{q\big(T(x_0)\big)} = \rho \big(\phi_d(x_0)\big)\frac{\big|\det\nabla T(x_0)\big|}{\pprior(x_0)}.
\end{equation}
Moreover,
\begin{equation}
\label{eq:Z-IS}
Z = \E_{x_0\sim\pprior}\big[w(x_0)\big], \qquad  \widehat Z = \frac1M\sum_{i=1}^M w\big(x_0^{(i)}\big)
\;\;\text{is an unbiased estimator of }Z.
\end{equation}

\begin{proof}
Because $T$ is a diffeomorphism, the change of variables formula gives 
\[
q(y)=\pprior\big(T^{-1}(y)\big)|\det\nabla T(T^{-1}(y))|^{-1}.
\]
Substituting $y=T(x_0)$ yields 
\[
q(T(x_0))=\frac{\pprior(x_0)}{|\det\nabla T(x_0)|},
\] which implies \cref{eq:det-flow-weight}. 
Then
\[
Z=\int\rho(y)\deriv y =\int \rho(T(x))|\det\nabla T(x)|\deriv x =\E_{x_0\sim\pprior}\bigg[\frac{\rho(T(x_0))|\det\nabla T(x_0)|}{\pprior(x_0)}\bigg],
\]
which is \cref{eq:Z-IS}. 
Unbiasedness of \(\widehat Z\) follows by linearity of expectation.
\end{proof}

\subsection{Hutchinson trace estimator}\label{sec:hutchison}
Evaluating $\nabla \cdot b_\theta(\rvx_t,t)=\mathrm{tr}\big(\nabla_x b_\theta(\rvx_t,t)\big)$ naïvely is $O(d^2)$ and requires forming Jacobians explicitly.
Following \citet{grathwohl2019ffjord}, we estimate the trace in linear time using the
Hutchinson identity:
\[
\mathrm{tr}(A)=\E_{\veps}\big[\veps^\top A \veps\big],
\]
for any square matrix $A$ and zero-mean probe $\veps$ with $\mathrm{Cov}(\veps)=I$ (e.g. Rademacher or standard Gaussian).
Applied to $A=\nabla_x b_\theta(\rvx_t,t)$, this gives the unbiased estimator
\[
\nabla \cdot b_\theta(\rvx_t,t) \approx \veps^\top \big(\nabla_x b_\theta(\rvx_t,t)\big) \veps,
\]
which can be computed with a pair of vector-Jacobian products (VJPs) at roughly the cost of one forward/backward evaluation.

\begin{figure}[t]
\centering
\includegraphics[width=0.60\textwidth]{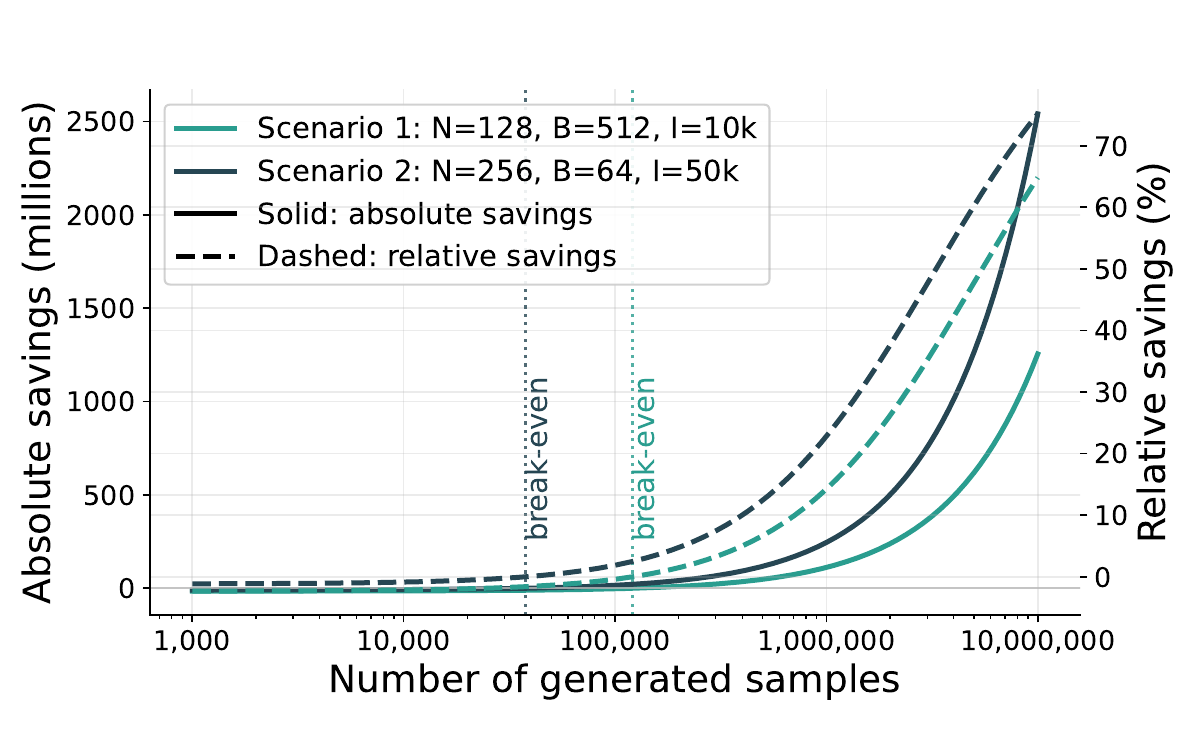}
\caption{Absolute and relative NFE savings of OSDS over the baseline diffusion samplers PIS, DDS as a function of the number of generated samples (log scale).}
\label{fig:nfe-tradeoff}
\end{figure}

\section{Cost-benefit analysis of one-step inference}\label{sec:cost-analysis}
A practical question is when the additional NFEs spent on the state- and volume-consistency branches pay off at inference time. 
Let $N$ denote the number of Euler-Maruyama steps used by a baseline diffusion sampler, $B$ the training batch size and $I$ the number of training iterations. 
A baseline method uses
\(
\mathrm{NFE}_{\text{base}}(S) = IBN + SN
\)
network evaluations to train and then generate $S$ samples. 
OSDS reuses the same $N$-step discretization for the FB-RND base loss but adds three extra evaluations per iteration for the teacher-student branches, and it applies a single PF-ODE step per test-time sample. 
Its total cost is
\(
\mathrm{NFE}_{\text{OSDS}}(S) = IB(N{+}3) + S.
\)
The total NFE savings after drawing $S$ samples are therefore
\(
\Delta_{\text{NFE}}(S)
= \mathrm{NFE}_{\text{base}}(S) - \mathrm{NFE}_{\text{OSDS}}(S)
= (N-1)S - 3IB,
\)
so OSDS becomes strictly cheaper once
\(
S > S_{\text{break}} := 3IB/(N-1).
\)
For $S \gg S_{\text{break}}$ the training term becomes negligible as the relative savings approach $1 - 1/N$.

\Cref{fig:nfe-tradeoff} illustrates $\Delta_{\text{NFE}}(S)$ and the corresponding relative savings for two representative configurations. 
Scenario~1 uses $N{=}128$, $B{=}512$, and $I{=}10\mathrm{k}$ iterations, corresponding to a moderately sized run.
Scenario~2 uses a more expensive baseline with $N{=}256$, $B{=}64$, and $I{=}50\mathrm{k}$.
These plots highlight that the distillation overhead is a one-time cost: once the sampler is reused for thousands to millions of draws, OSDS yields substantial end-to-end computational savings.

\section{Implementation Details}
\subsection{Model architecture and optimization}\label{sec:model-opt}
All methods use a \textsc{PISGRAD} backbone of identical size (same depth/width and Fourier features for time) \citep{zhang2022pis}.
Our variant (step-conditioned) augments the backbone with a small embedding for the step size $d$:
we compute Fourier features for both $t$ and $d$, concatenate them, and feed the result into the time coder.

We use AdamW \citep{loshchilov2017adamW} with decoupled weight decay~$0.1$ and global-norm gradient clipping (threshold $1.0$ by default).
All implementations are in JAX/Flax/Optax \citep{jax2018github}. 
We maintain an EMA of parameters with decay $0.999$.
We train and evaluate using a NVIDIA A6000 GPU.

The learning rate is tuned for all methods in $\{0.001, 0.0001, 0.00001\}$, following \citet{blessing2024elbos}.
We use the model checkpoints with the best moving average ELBO.

\subsubsection{Jacobian regularization}\label{sec:jac-reg}
To stabilize the deterministic flow during distillation, we add a small Jacobian-norm penalty on the step-conditioned field used by the student.
Let $J_u(x,t,d)=\partial u_\theta(x,t,d)/\partial x \in \mathbb{R}^{d\times d}$ denote the Jacobian w.r.t.\ the state.
We define
\begin{equation}
  L_{\mathrm{Jac}}(x_t; t, d) = \mathbb{E}_{v}\Bigl[\frac{1}{d}\|J_u(x_t,t,d)v\|_2^2\Bigr] \approx \frac{1}{d}\|J_u(x_t,t,d)v\|_2^2,
\end{equation}
where $v$ is a single Rademacher probe ($v_i \sim \mathrm{Unif}\{\pm1\}$). 
This is a Hutchinson-style estimator:
$\mathbb{E}_v\|Jv\|_2^2=\mathrm{tr}(J^\top J)=\|J\|_F^2$, so $L_{\mathrm{Jac}}$ penalizes the (scaled) Frobenius norm of the Jacobian and hence the local Lipschitz constant of $u_\theta$ (we absorb the $1/d$ scaling into the weight).
We compute $J_u(x_t,t,d) v$ with a single JVP at the start of the student's step $x_t$.

The full distillation objective is then
\[
  L = L_{\text{state}} + \lambda_{\text{trace}}L_{\text{trace}} + \lambda_{\mathrm{Jac}}L_{\mathrm{Jac}},
\]
This regularization mirrors the JVP/trace machinery used in continuous/ODE flows: Hutchinson probes are standard for Jacobian traces in, and controlling Lipschitzness improves numerical stability and reduces stiffness (which otherwise increases solver work). 
See the instantaneous change-of-variables and trace/JVP discussion in CNFs \citep{chen2018ctnf} and FFJORD \citep{grathwohl2019ffjord}.

\subsection{Loss algorithmic details}
This section spells out how we instantiate the three computations used by OSDS: (i) the fine‑resolution forward–backward Radon-Nikodym (RND) training loss, (ii) the teacher-student self‑distillation losses (state and volume), and (iii) the discrete forward–backward (FB–RND) bound used for evaluation when a reliable backward kernel is available.

Algorithm~\ref{alg:fb-rnd-bound} evaluates the discrete FB likelihood ratio (\cref{prop:disc-fb}, \cref{eq:disc-fb-rnd-main,eq:disc-fb-bound-main}).
Algorithm~\ref{alg:distill} implements the shortcut self‑distillation objective (\cref{eq:state-consistency-loss,eq:vol-loss,eq:final-loss}).

The self-distillation loss adds a minimal computation overhead of three additional NFEs per iteration (out of $128$, $256$ or $512$, for instance). 

\begin{algorithm}[t]
\caption{Discrete forward–backward sampling ELBO/FB‑RND (\cref{prop:disc-fb})}
\label{alg:fb-rnd-bound}
\begin{algorithmic}[1]
\REQUIRE trained $\theta$; densities $\rho$, $\pi$; steps $N$; transition densities $p_{n+1|n}$ and time‑adjoint $p_{n|n+1}$ defined by the discrete chain.
\ENSURE path $x_{0:N}$, path weight $w$, and ELBO contribution $\log w$.
\STATE $x_0 \sim \pi$
\STATE $x_n \gets x_0$
\STATE $\log w \gets 0$
\FOR{$n=0$ {\bf to} $N-1$}
    \STATE $x_{n+1} \sim p_{n+1|n}(x_{n+1}\!\mid x_n)$
    \STATE $\log w \gets \log w + \log p_{n+1|n}(x_{n+1}\!\mid x_n) - \log p_{n|n+1}(x_n\!\mid x_{n+1})$
    \STATE $x_n \gets x_{n+1}$
\ENDFOR
\STATE $x \gets x_{n}$
\STATE $\log w \gets \log w + \log \rho(x_N) - \log \pi(x_0)$
\STATE \textbf{return} $x_{0:N}$, $w = e^{\log w}$, $\log w$
\end{algorithmic}
\end{algorithm}

\begin{algorithm}[t]
\caption{Self-distillation losses}
\label{alg:distill}
\begin{algorithmic}[1]
\REQUIRE Anchors $\{(x_t^{(b)},t)\}_{b=1}^B$; per-anchor steps $\{d^{(b)}\}_{b=1}^B$; control $u_\theta$; PF ODE drift $b_\theta(x,\tau,d)$; integrator $\Psi$ that can integrate an augmented ODE $(\dot x,\dot\ell)$ and return both final state and log-volume; frozen teacher params $\theta'=\texttt{stopgrad}(\theta)$; shared Hutchinson probe $\varepsilon$.
\ENSURE $\mathcal{L}_{\text{state}}$, $\mathcal{L}_{\text{vol}}$.
\STATE $\text{sum}_{\text{state}}\leftarrow 0$;\quad $\text{sum}_{\text{vol}}\leftarrow 0$
\FOR{$b=1$ \textbf{to} $B$}
  \STATE $x \leftarrow x_t^{(b)}$;\quad $t_0 \leftarrow t$;\quad $d \leftarrow d^{(b)}$
  \STATE Define teacher augmented drift (half-step):
  \STATE \hspace{1em}$g_{\theta'}^{(d/2)}(x,\ell,\tau) \;=\; \big(b_{\theta'}(x,\tau,\tfrac d2), \textsc{Hutchinson}(\nabla_x b_{\theta'}(x,\tau,\tfrac d2),\,\varepsilon)\big)$
  \STATE Teacher half-step 1: $(\tilde x,\tilde v) \leftarrow \Psi\big(g_{\theta'}^{(d/2)}, (x,0), t_0, \tfrac d2\big)$
  \STATE Teacher half-step 2: $(x^{\text{teach}}_{t+d},\bar v) \leftarrow \Psi\big(g_{\theta'}^{(d/2)}, (\tilde x,0), t_0+\tfrac d2, \tfrac d2\big)$
  \STATE $v_{\text{teach}} \leftarrow \tilde v + \bar v$
  \STATE Define student augmented drift (full-step):
  \STATE \hspace{1em}$g_{\theta}^{(d)}(x,\ell,\tau) = \big(b_{\theta}(x,\tau,d), \textsc{Hutchinson}(\nabla_x b_{\theta}(x,\tau,d),\varepsilon)\big)$
  \STATE Student single step: $(\widehat x_{t+d}, v_{\text{stud}}) \leftarrow \Psi\big(g_{\theta}^{(d)}, (x,0), t_0, d\big)$
  \STATE $\text{sum}_{\text{state}} \leftarrow \text{sum}_{\text{state}} + \|\widehat x_{t+d} - x^{\text{teach}}_{t+d}\|^2$
  \STATE $\text{sum}_{\text{vol}} \leftarrow \text{sum}_{\text{vol}} + (v_{\text{stud}} - v_{\text{teach}})^2$
\ENDFOR
\STATE $\mathcal{L}_{\text{state}} \leftarrow \text{sum}_{\text{state}}/B$;\quad $\mathcal{L}_{\text{vol}} \leftarrow \text{sum}_{\text{vol}}/B$
\STATE \textbf{return} $\mathcal{L}_{\text{state}},\,\mathcal{L}_{\text{vol}}$
\end{algorithmic}
\end{algorithm}

\subsection{VP SDE and time schedule}\label{sec:vp-sde}
Unless specified otherwise, we use the variance–preserving (VP) noising SDE
\citep{ho2020ddpm, song2021sde}
\begin{equation}
  \deriv \rvx_t = -\frac12\beta(t)\rvx_t \deriv t + \sqrt{\beta(t)} \sigma_0 \deriv \rvw_t, \qquad t\in[0,1],
  \label{eq:vp-noising}
\end{equation}
where \(\sigma_0>0\) is a constant scale and \(\beta(t)\) is a nonnegative schedule. 
In our experiments we use a linear schedule $\beta(t) = \beta_{\text{min}} + t(\beta_{\text{max}} - \beta_{\text{min}})$ with $\beta_{\text{min}} = 0.01$ and $\beta_{\text{max}} = 10$.

For all methods we use the same cosine schedule for the noise scale.
For $N$ total steps and index $k\in\{0,\dots,N\}$, define $t_k := k/N$.
With hyperparameters $\sigma_{\max} >\sigma_{\min} > 0$, $s \ge 0$, and exponent $p \ge 1$, we use
\begin{equation}\label{eq:cosine-sched}
  \sigma(t_k) = \frac12\bigl(\sigma_{\max}-\sigma_{\min}\bigr)
  \cos^p \Bigl(\frac{\pi}{2} \frac{1+s - t_k}{1+s}\Bigr) + \frac12 \sigma_{\min}.
\end{equation}

\section*{Limitations}
Our work has several limitations.
First, our empirical study focuses on synthetic energy landscapes and low- to mid-dimensional Bayesian benchmarks (up to around $60$ dimensions). 
We do not include molecular systems such as Lennard-Jones clusters or alanine dipeptide, which are standard in recent work on Boltzmann sampling. 

Second, OSDS is built on top of a diffusion-based PIS-GRAD backbone trained at a relatively fine time discretization. 
Our method amortizes this training into one- or few-step inference, but it does not reduce the cost of learning the teacher itself, and we do not explore alternative backbones or simulation-free training.

Finally, our deterministic-flow importance weights rely on estimating log-Jacobian divergences via Hutchinson-style trace estimators along the PF ODE. 
While our experiments suggest that this is numerically stable in the dimensionalities we consider, we do not provide a systematic study of variance and bias as the state dimension grows.

\end{document}